\documentclass[letterpaper, 10 pt, journal, twoside]{IEEEtran} 

\IEEEoverridecommandlockouts                              % This command is only needed if 
\usepackage[l2tabu,orthodox]{nag}

\usepackage[
    backend=bibtex,
    style=ieee,
    sorting=none,
    natbib=true,
    doi=false,
    isbn=false,
    url=false,
    eprint=false,
    maxcitenames=1,
    mincitenames=1
]{biblatex}

\usepackage[pdftex,colorlinks]{hyperref}
\usepackage[printonlyused]{acronym}

\usepackage{siunitx}
\usepackage[all]{nowidow}

\usepackage{tikz}
\usetikzlibrary{arrows}
\usetikzlibrary{arrows.meta}
\usepackage{makecell}
\usetikzlibrary{positioning}
\usetikzlibrary{decorations.pathreplacing,calligraphy}

\usepackage{lipsum}

\usepackage{hhline}

\usepackage{xspace} %smart handling of space in commands

\usepackage{epstopdf}

\usepackage{import}

\usepackage{booktabs}

\usepackage{tabularx}
\usepackage{multirow, multicol}

\newlength{\Oldarrayrulewidth}
\newcommand{\Cline}[2]{%
  \noalign{\global\setlength{\Oldarrayrulewidth}{\arrayrulewidth}}%
  \noalign{\global\setlength{\arrayrulewidth}{#1}}\cline{#2}%
  \noalign{\global\setlength{\arrayrulewidth}{\Oldarrayrulewidth}}}
\newcolumntype{?}[1]{!{\vrule width #1}}

\usepackage{amssymb,amsfonts,amsmath,amscd}

\usepackage{bm}

\usepackage{balance}
\usepackage[T1]{fontenc}

\newcommand{\rev}[1]{{\color{black}{#1}}}
\usepackage{cancel}

\newcommand{\bbm}{\begin{bmatrix}}
\newcommand{\ebm}{\end{bmatrix}}

\newcommand{\ignore}[1]{}

\newcommand{\bma}[1]{\left[\begin{array}{#1}}
\newcommand{\ema}{\end{array}\right]}

\DeclareMathAlphabet{\mbf}{OT1}{ptm}{b}{n}
\newcommand{\mbs}[1]{{\boldsymbol{#1}}}
\def\fdotb{{\raisebox{-0.6ex}{ \kern0.2ex\raisebox{0.8ex}{\tiny $\hspace*{-1ex}\circ$}}}}
\def\fddotb{{\raisebox{-0.6ex}{ \kern0.2ex\raisebox{0.8ex}{\tiny $\hspace*{-1ex}\circ\circ$}}}}

\newcommand{\f}{\frac}

\newcommand{\trans}{{\ensuremath{\mathsf{T}}}} % transpose
\newcommand{\utimes}{ {\raisebox{-0.6ex}{ \kern-1.0ex\raisebox{0.6ex}{ \small $\mathsf{v}$}}} } % 
\newcommand{\beq}{\begin{equation}}
\newcommand{\eeq}{\end{equation}}
\newcommand{\bdis}{\begin{displaymath}}
\newcommand{\edis}{\end{displaymath}}
\newcommand{\beqarray}{\begin{eqnarray}}
\newcommand{\eeqarray}{\end{eqnarray}}
\newcommand{\beqarraynn}{\begin{eqnarray*}}
\newcommand{\eeqarraynn}{\end{eqnarray*}}

\DeclareMathAlphabet{\mbf}{OT1}{ptm}{b}{n}

\newcommand{\diag}{{\ensuremath{\mathrm{diag}}}}

\title{Are Doppler Velocity Measurements Useful for Spinning Radar Odometry?}

\author{Daniil Lisus$^{1}$, Keenan Burnett$^{1}$, David J. Yoon$^{1}$, Richard Poulton$^{2}$, John Marshall$^{2}$, and Timothy D. Barfoot$^{1}$% <-this % stops a space
\thanks{Manuscript received: July, 14, 2024; Revised October, 10, 2024; Accepted November, 11, 2024.}
\thanks{This paper was recommended for publication by Editor 
Sven Behnke upon evaluation of the Associate Editor and Reviewers' comments.
This work was supported by the Ontario Graduate Scholarship (OGS) Program provided by the Province of Ontario.}
\thanks{$^{1}$ Authors are with the University of Toronto Institute for Aerospace Studies (UTIAS), 4925 Dufferin St, Ontario, Canada. {\tt\footnotesize daniil.lisus@robotics.utias.utoronto.ca}}
\thanks{$^{2}$ Authors are with Navtech Radar, Home Farm, School Rd, Ardington, Wantage, Oxfordshire, UK. {\tt\footnotesize{richard.poulton@navtechradar.com}}}% <-this % stops a space
\thanks{Digital Object Identifier (DOI): see top of this page.}
}

\begin{document}

\newpage
%auto-ignore
% This is not a standalone latex document. To use this file
% as a cover page on an arXiv upload of a document that is 
% already accepted as some sort of IEEE publication, you must
%
%  1) add the following just after the \begin{document} line
%     of your main paper document
%
%         \input{arxiv-cover-ieee.tex}
%
%  2) and replace the relevant information in the block below.
%
% The relevant information has been parameterized as variables.
% Simply replace the variable values with your stuff and the 
% result should be good.
%
% Make sure to not include this file for ACTUAL submissions to 
% the IEEE. Luckily you can just comment in/out the 
% \input{arxiv-cover-ieee.tex} line.
%
% FYI: The exact citation with formatting can be obtained 
% from your paper's page on IEEE Xplore.
%
%%%%%%%%%%%%%%%%%%%%%%%%%%%%%%%%%%%%%%%%%%%%%%%%%%%%%%%%%%%%%%%
%%%%%%%%%%%%%%%%%%%%%% ADD YOUR INFO HERE %%%%%%%%%%%%%%%%%%%%%
%%%%%%%%%%%%%%%%%%%%%%%%%%%%%%%%%%%%%%%%%%%%%%%%%%%%%%%%%%%%%%%
\def \myJournal {IEEE Robotics and Automation Letters}
\def \myDoi {10.1109/LRA.2024.3505821}
\def \myPaperSiteName {IEEE Xplore}
\def \myPaperSiteLink {https://ieeexplore.ieee.org/document/10766427}
\def \myYear {2024}

\def \myPaperCitation{D. Lisus, K. Burnett, D. J. Yoon, R. Poulton, J. Marshall and T. D. Barfoot, ``Are Doppler Velocity Measurements Useful for Spinning Radar Odometry?,'' in \textit{IEEE Robotics and Automation Letters}, vol. 10, no. 1, pp. 224-231, Jan. 2025.}

%%%%%%%%%%%%%%%%%%%%%%%%%%%%%%%%%%%%%%%%%%%%%%%%%%%%%%%%%%%%%%%
%%%%%%%%%%%%%%%%%%%%%%%%%%%%%%%%%%%%%%%%%%%%%%%%%%%%%%%%%%%%%%%

\begin{figure*}[t]

\thispagestyle{empty}
\begin{center}
\begin{minipage}{6in}
\centering
This paper has been accepted for publication in \emph{\myJournal}. 
\vspace{1em}

This is the author's version of an article that has, or will be, published in this journal or conference. Changes were, or will be, made to this version by the publisher prior to publication.
\vspace{2em}

\begin{tabular}{rl}
DOI: & \myDoi\\
\myPaperSiteName: & \texttt{\myPaperSiteLink}
\end{tabular}

\vspace{2em}
Please cite this paper as:

\myPaperCitation

\vspace{15cm}
\copyright \myYear \hspace{4pt}IEEE. Personal use of this material is permitted. Permission from IEEE must be obtained for all other uses, in any current or future media, including reprinting/republishing this material for advertising or promotional purposes, creating new collective works, for resale or redistribution to servers or lists, or reuse of any copyrighted component of this work in other works.

\end{minipage}
\end{center}
\end{figure*}
\newpage
\clearpage
\pagenumbering{arabic} 

\maketitle

%%%%%%%%%%%%%%%%%%%%%%%%%%%%%%%%%%%%%%%%%%%%%%%%%%%%%%%%%%%%%%%%%%%%%%%%%%%%%%%%
\begin{abstract}
Spinning, frequency-modulated continuous-wave (FMCW) radars with $360 \si{\degree}$ coverage have been gaining popularity for autonomous-vehicle navigation. However, unlike `fixed' automotive radar, commercially available spinning radar systems typically do not produce radial velocities due to the lack of repeated measurements in the same direction and the fundamental hardware setup. To make these radial velocities observable, we modified the firmware of a commercial spinning radar to use triangular frequency modulation. In this paper, we develop a novel way to use this modulation to extract radial Doppler velocity measurements from \rev{consecutive azimuths of a radar intensity scan,} without any data association. We show that these noisy, error-prone measurements contain enough information to provide good ego-velocity estimates, and incorporate these estimates into different modern odometry pipelines. We extensively evaluate the pipelines on over $110 \, \si{\km}$ of driving data in progressively more geometrically challenging autonomous-driving environments. We show that Doppler velocity measurements improve odometry in well-defined geometric conditions and enable it to continue functioning even in severely geometrically degenerate environments, such as long tunnels.
\end{abstract}


\section{Introduction}
    \label{sec:introduction}
    \IEEEPARstart{R}{ange}-measuring sensors, such as radar and lidar, are now commonly used on autonomous vehicles (AVs) for tasks such as adaptive cruise control, crash warning systems, odometry, and localization. Lidar is generally more accurate than radar owing to its ability to capture returns from the full 3D environment at a high rate and accuracy \cite{survey_lidar_loc}. Radar tends to have a longer range and is essentially unaffected by precipitation, fog, smoke, and sensor cleanliness. Radar has also traditionally been the only sensor capable of using the Doppler effect to estimate relative radial velocities of objects. As a result, there has been a recent increase in work attempting to leverage the unique advantages of radar \cite{a_new_wave_radar}.

\begin{figure}[t]
    \centering
    \scalebox{0.93}{
    \begin{tikzpicture}    
        \node (vel_plot) at (0,4.2) {\includegraphics[width=0.49\textwidth, trim={0 0cm 0 0cm},clip]{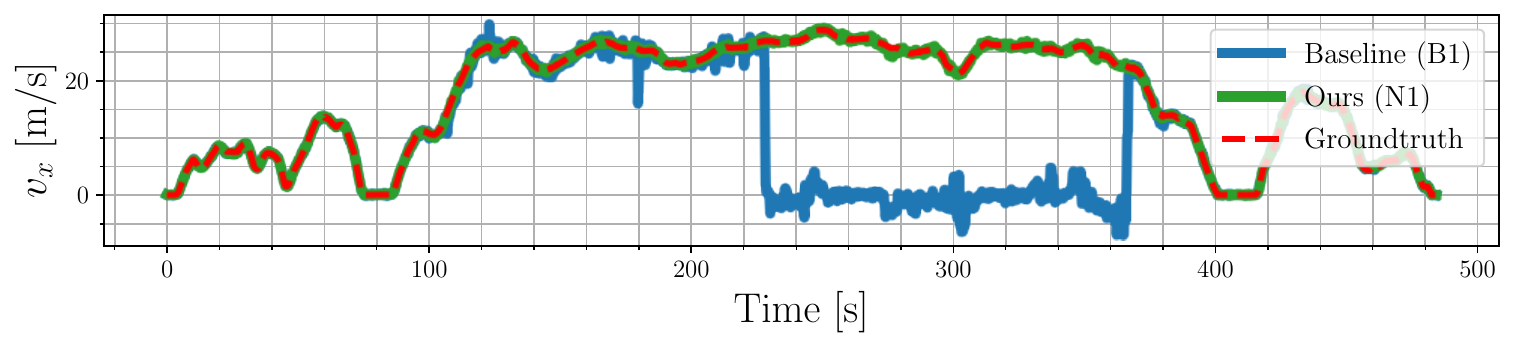}};
    
        \node (photo) at (0,0) {\includegraphics[width=0.48\textwidth, trim={0 0cm 0 0cm},clip]{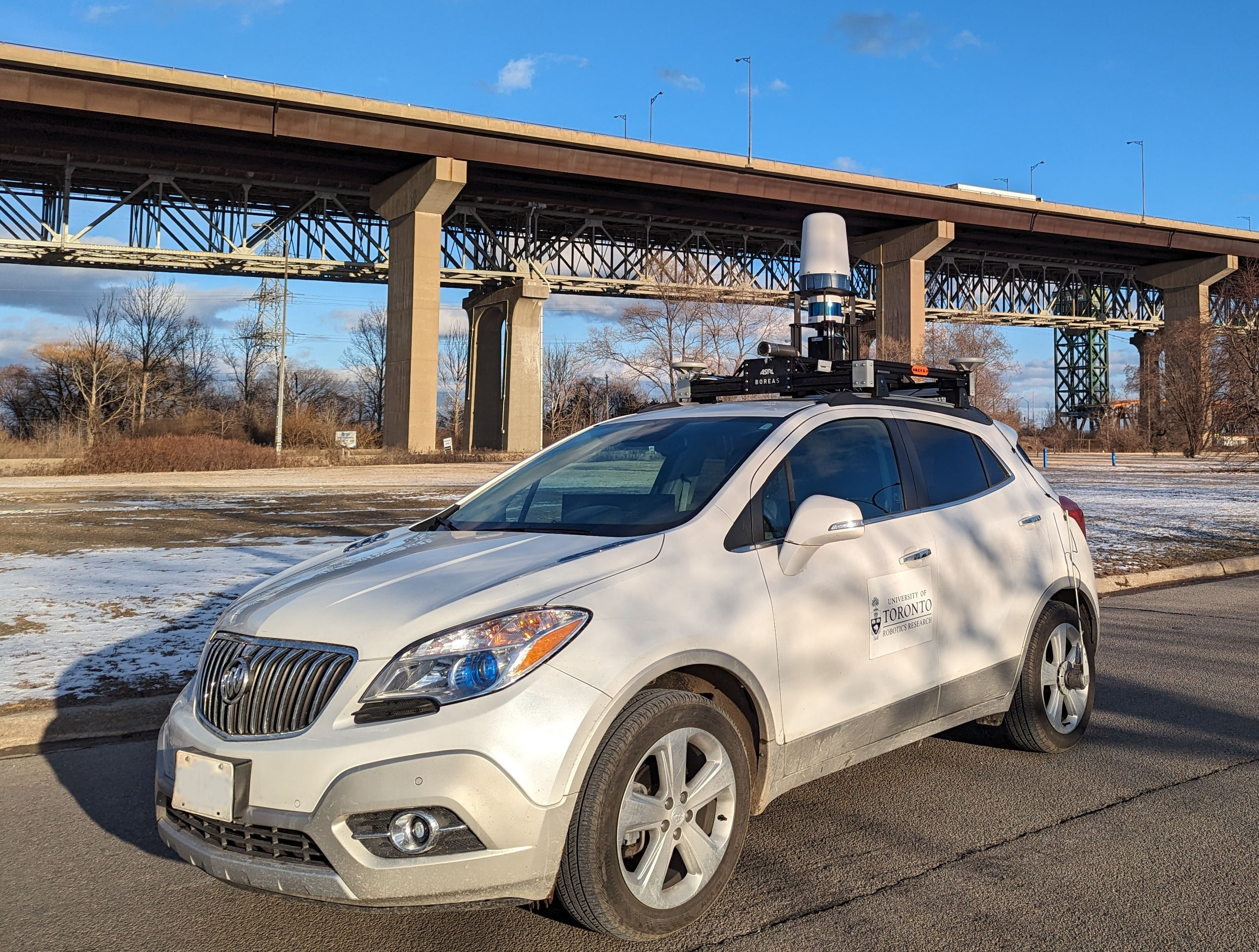}};
        % GNSS & gyro
        \node[draw=black, rounded corners, fill=white, inner sep=5pt] (GNSS) at (-1,0.85) {\small \textbf{GNSS/IMU}};
        \draw[-{Triangle[length=1.8mm,width=2.8mm]}, red, line width=0.7mm] (GNSS) -- (1.22,0.85);
        % radar
        \node[draw=black, rounded corners, fill=white, inner sep=5pt] (radar) at (3.25,0.85) {\small \textbf{Radar}};
        \draw[-{Triangle[length=1.8mm,width=2.8mm]}, red, line width=0.7mm] (radar) -- (3.25,1.5) -- (1.48,1.5);
        % lidar
        \node[draw=black, rounded corners, fill=white, inner sep=5pt] (lidar) at (-1,1.5) {\small \textbf{Lidar}};
        \draw[-{Triangle[length=1.8mm,width=2.8mm]}, red, line width=0.7mm] (lidar) -- (0.4,1.5) -- (0.4,1.15) -- (1.22,1.15);
        % skyway
        \node[draw=black, rounded corners, fill=white, inner sep=5pt] (skyway) at (3.25,2.7) {\small \textbf{Skyway}};
        \draw[-{Triangle[length=1.8mm,width=2.8mm]}, red, line width=0.7mm] (skyway) -- (1.6, 2.7) -- (1.6, 2.05);
        \draw[-{Triangle[length=1.8mm,width=2.8mm]}, red, line width=0.7mm] (skyway) -- (1.6, 2.7) -- (1.6, 3.65);

        \draw[very thick,decorate,decoration={brace,amplitude=10pt,aspect=0.2},red] (2.5,4) -- (-2,4);
    \end{tikzpicture}
    }
    \caption{\rev{Top: The estimated vehicle velocity using a baseline and our Doppler-enabled method.
    Bottom: Our data collection platform Boreas in front of the Burlington Bay James N. Allan Skyway. The skyway has few reliable geometric features, resulting in the baseline estimation failure.}}
    \label{fig:boreas}
\end{figure}

AVs typically use frequency-modulated continuous-wave (FMCW) phased-array or spinning radars. Phased-array radars directly provide the range, elevation, radial cross section, and radial velocity of targets, but only operate on a limited field-of-view (FOV) \cite{Gao_Zhang_Wang_Lu}. In contrast, spinning radars provide a full $360\si{\degree}$ FOV, but typically only return the range and intensity of targets and not their relative velocities \cite{boreas}. Spinning radars are preferred for navigation tasks such as odometry and localization, where a high degree of accuracy and awareness of all vehicles on the road is sought \cite{Kellner_2013}. Using multiple phased-array radars for these tasks would require several sensors and extensive calibration. Additionally, most automotive phased-array radars directly provide extracted features instead of raw signal returns. This makes it harder to fine-tune the feature extraction process for specific tasks.

To expand the capabilities of spinning radar, we adopted a `triangular' frequency modulation scheme to allow extraction of Doppler velocity measurements.
\rev{This scheme, as opposed to the typically used `sawtooth' modulation, requires no hardware changes and does not affect the range or angular resolution.}
Our Doppler-enabled radar system, which we refer to as `spinning Doppler radar', alternates its modulation pattern per azimuth, first increasing and then decreasing the frequency between consecutive azimuths.
This makes it theoretically possible to extract the Doppler velocity of a target.
However, in contrast to automotive radar, which provides accurate Doppler measurements from many repeated measurements of the same target \cite{4404963}, spinning radar typically has only a $50\%$ overlap between consecutive azimuths.
This means that a target is present in only two to three measurements per rotation at best.
Thus, despite triangular modulation making it possible to extract Doppler velocities from spinning radar, it is unclear if they would be usable.
This paper aims to answer the question: are Doppler measurements useful for spinning radar odometry?

Previously, \citet{Vivet_Checchin_Chapuis_2013} used Doppler measurements for radar localization and mapping using a custom spinning Doppler radar. More recently, \citet{Rennie_Williams_Newman_DeMartini_2023} made use of the Doppler information from a Navtech RAS6 radar to train a neural network to directly predict vehicle pose information. However, they did not directly extract the Doppler velocity information, which makes it hard to evaluate its quality and impact. \rev{We are thus revisiting the analytical approach presented by \citet{Vivet_Checchin_Chapuis_2013} and refining it using modern approaches. Additionally, both previous works halved the number of azimuths in order to extract Doppler information, and, in part because of this, do not produce results that are on par with modern state-of-the-art radar odometry approaches. Our approach extracts Doppler velocities while retaining the full angular resolution. We evaluate the usefulness of our extracted Doppler measurements on a dataset collected in progressively more challenging environments ranging from a suburb to the geometrically degenerate skyway shown in Figure \ref{fig:boreas}.}

The main contributions of this paper are as follows:
\rev{
\begin{enumerate}
    \item We present a novel approach to extracting Doppler velocities from spinning radar that leverages filtering, a robust cost function, and bias estimation to improve upon prior work by \citet{Vivet_Checchin_Chapuis_2013}.
    \item We demonstrate the utility of these Doppler velocity measurements by performing radar odometry experiments on over $110 \, \si{\km}$ of driving data.
\end{enumerate}
}

The rest of the paper is organized as follows. Section \ref{sec:RW} contextualizes the novelty claim. Section \ref{sec:methodology} discusses relevant radar theory and our Doppler velocity extraction approach. Section \ref{sec:experiments} presents the dataset, pipelines, and results. Finally, Section \ref{sec:conclusion} concludes the paper.
\section{Related Work}
    \label{sec:RW}
    \subsection{Spinning Radar Navigation}
Spinning radar navigation has recently seen a surge of interest \cite{a_new_wave_radar, radar_survey_2, venon2022millimeter}.
Odometry and localization methods have typically been split into iterative closest point (ICP)-style pointcloud matching approaches \cite{Cen_filtering, cen2019, are_we_ready_for, Adolfsson_Magnusson_Alhashimi_Lilienthal_Andreasson_2023, 10610311}, scan matching approaches \cite{masking_by_moving, Checchin_Gérossier_Blanc_Chapuis_Trassoudaine_2010}, and feature matching approaches \cite{under_the_radar, hero_paper, 10160997}. These have largely ignored Doppler-based distortion, and, owing to the limitations of previously available hardware, have not extracted Doppler measurements directly. \citet{do_we_need_to_compensate} corrected Doppler distortion in a spinning radar through the use of the vehicle ego-velocity estimate. 

\subsection{Automotive Doppler Radar}
\citet{Kellner_2013} showed the ability of an automotive radar sensor to produce usable ego-motion estimates in combination with an Ackerman vehicle model.
They did this by finding a least-squares fit to the measured point-wise radial Doppler velocities from points deemed stationary by Random Sample and Consensus (RANSAC) \cite{RANSAC}.
They showed that the ego-motion estimation fails in the presence of many non-stationary targets, such as can be expected in heavy road traffic.
\citet{Kellner_2014} extended this work by using multiple radars to relax the requirement for a vehicle model;
this is the first paper to show full 2D vehicle ego-motion estimation using only radar. 
More recently, \citet{Luque_Kubelka_Magnusson_Ruiz} showed that Doppler measurements from an automotive radar can be used to do 3D odometry by using RANSAC and a kinematic model.
Instead of using RANSAC, \citet{Michaelis_Berthold_Luettel_Wuensche} developed a dynamic object outlier rejection scheme using the previous ego-vehicle estimate. \citet{Gao_Zhang_Wang_Lu} used Doppler measurements to undistort radar points in an uncertainty-aware way before using them in a probabilistic descriptor-based localization approach. \citet{Kubelka_Fritz_Magnusson_2023} evaluated different automotive radar odometry pipelines and found that directly integrating high quality Doppler measurements with an IMU can outperform ICP-based methods in geometrically degenerate environments. \rev{\citet{10610311} proposed two Doppler-enabled odometry pipelines: a point-to-point (p2p) ICP one augmented with Doppler radial velocity terms, and a direct velocity estimation one based on \cite{Kellner_2013} with an added dynamic object-filtering step. Our ICP and Doppler pipeline is similar to theirs, although ours is implemented in continuous time with a white-noise-on-acceleration (WNOA) prior.}

\subsection{Spinning Doppler Radar}
Two previous works have used Doppler measurements from spinning radar. \citet{Vivet_Checchin_Chapuis_2013} developed a custom Doppler-enabled spinning radar and used it for the task of localization and mapping of a small mobile platform in simple outdoor environments. Our Doppler extraction approach is most similar to theirs, as both use cross-correlation. However, our extraction approach makes use of an additional filtering step\rev{, robust cost functions, and velocity bias correction}, all of which we found crucial to achieve velocity estimates of high enough quality to be useful in an autonomous-driving context. Additionally, for a radar scan composed of $N$ azimuths, our extractor produces $N-1$ velocity estimates as compared to their $N/2$. Finally, their experiments were conducted in relatively slow ($\leq 8.5 \, \si{\meter/\second}$) conditions without dynamic objects in the scene. We provide more extensive experimental results in multiple real autonomous-driving environments, with speeds reaching up to $28 \, \si{\meter/\second}$.

On the other hand, \citet{Rennie_Williams_Newman_DeMartini_2023} used a modern, commercially available Navtech RAS6 radar to estimate the vehicle ego-motion from the Doppler effect. This radar is functionally the same to the one that we use. However, their approach is entirely learning-based, as they feed their raw scans into a neural network to directly estimate the change in vehicle pose. They thus never directly extract Doppler measurements or estimate the ego-vehicle velocity explicitly. We propose a method to extract Doppler velocities, and show how they can be used in different ways to correct the ego-vehicle velocity, and, ultimately, position.

\subsection{Doppler Lidar}
It is also worth noting the development of FMCW lidar sensors, as they are also capable of providing Doppler measurements that have historically been unique to radar~\cite{pierrottet2008linear}. These sensors have thus far been limited to a fixed FOV, not unlike automotive radar. Doppler measurements from a commercially available FMCW lidar were shown to be effective at preventing ICP algorithm failure in degenerate geometric environments \cite{Hexsel_Vhavle_Chen_2022}. They were also shown to be useful in continuous-time odometry ICP \cite{picking_up_speed} and at fast, correspondence-free velocity estimation \cite{need_for_speed}. We evaluate similar pipelines using a spinning radar in our experiments.
\section{Methodology}
    \label{sec:methodology}
    \begin{figure}[t]
    \centering
    \includegraphics[width=0.43\textwidth]{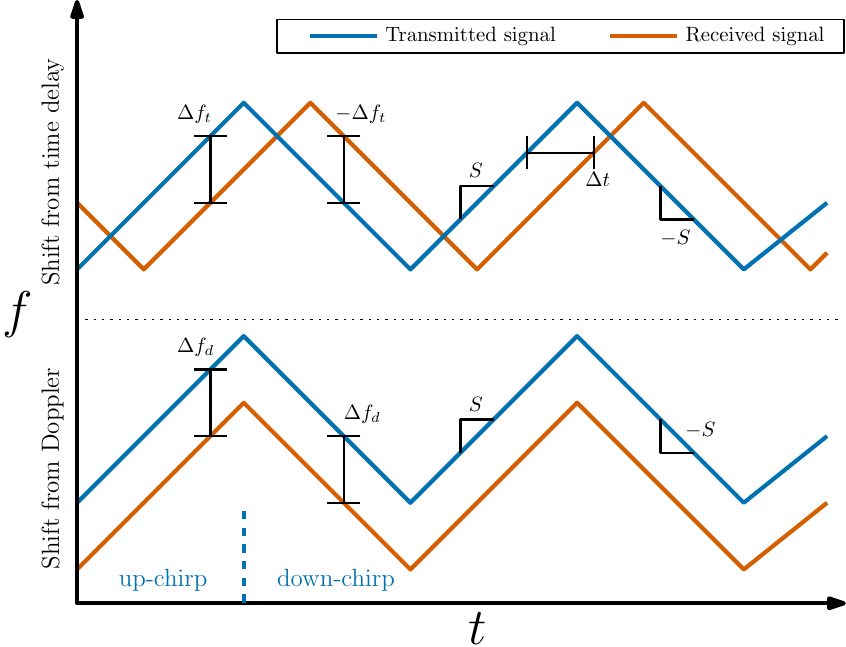}
    \caption{A graphical representation of the shift in the frequency between the transmitted signal in blue and the received signal in orange for an FMCW radar with a triangular modulation pattern of slope $S$. The top shows a frequency shift $\Delta f_t$ resulting from a time delay $\Delta t$ induced by the signal traveling some distance. The bottom shows a frequency shift $\Delta f_d$ resulting from the Doppler effect of an object moving away from the radar.}
    \label{fig:f_vs_t}
\end{figure}

\subsection{Spinning Doppler Radar Theory}
\begin{figure}[t]
    \centering
    \begin{tikzpicture}
        % Static subfigure
        % Fig background
        %trim={0 0 0 11.25cm}
        \node (static) at (0,0) {\includegraphics[width=0.48\textwidth, trim={3.5cm 5cm 6cm 14cm},clip]{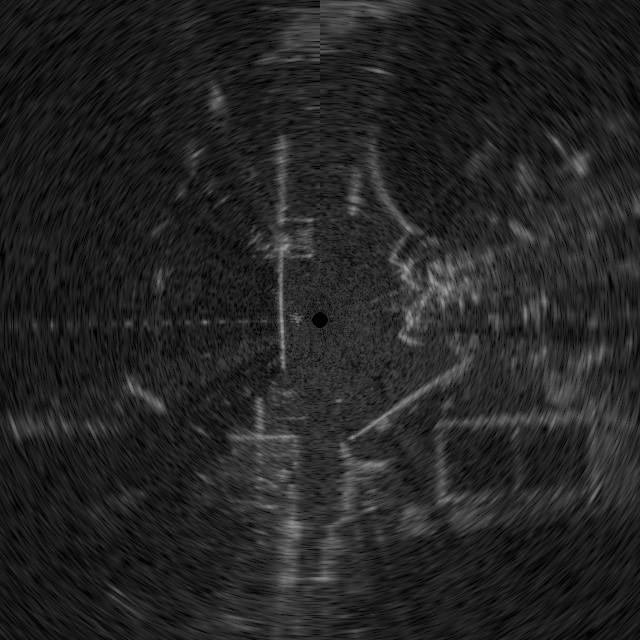}};
        % Signal label
        %(-2.6,-1.0)
        \node[draw=black, rounded corners, fill=white, inner sep=5pt] (top_signal) at (-2.6,0.0) {
            \begin{minipage}{0.13\textwidth}
                \centering
                \includegraphics[width=\linewidth]{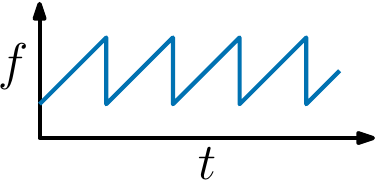} \\
                \textbf{Sawtooth}
            \end{minipage}
        };
        % Dynamic subfigure
        % (-2.6,-5.25)
        \node (dynamic) at (0,-2.30) {\includegraphics[width=0.48\textwidth, trim={3.5cm 5cm 6cm 14cm},clip]{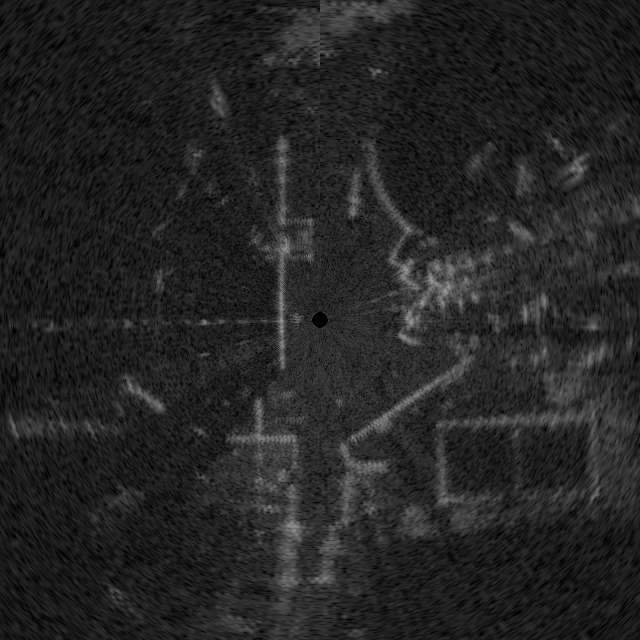}};
        \node[draw=black, rounded corners, fill=white, inner sep=5pt] (top_signal) at (-2.6,-2.30) {
            \begin{minipage}{0.13\textwidth}
                \centering
                \includegraphics[width=\linewidth]{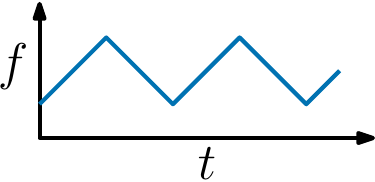} \\
                \textbf{Triangular}
            \end{minipage}
        };
        % Separation line
        \draw[-, black, line width=1.5pt](-4.267,-1.15) -- (4.267,-1.15);
    \end{tikzpicture}
    \caption{Static features measured using a sawtooth (top) and triangular (bottom) modulated moving radar. The alternating Doppler-induced shift produced by the triangular signal modulation pattern can be seen as a `zig-zag' in the intensity returns of continuous, flat features in the bottom image.}
    \label{fig:normal_vs_doppler_scan}
\end{figure}

FMCW radar determines the range $r$ to a target based on the difference between the received and transmitted signal frequencies $\Delta f$:
\begin{align}\label{eq:r_basic}
    r = \frac{c \, \Delta f}{2 S},
\end{align}
where $c$ is the speed of light, $S$ is the slope of the modulation pattern, and the division by 2 is present since the signal has to travel to and from the target.
The difference in frequencies arises from two factors: the time delay $\Delta t$ corresponding to the signal travelling to and from an object, and a Doppler-induced compression or expansion of the signal from the relative motion of a target.
These two cases are visualized in Figure \ref{fig:f_vs_t} for the case of a triangular modulation signal, where $S$ is first positive (`up chirp') and then negative (`down chirp') in alternating fashion.
When only a time-induced shift $\Delta f_t$ is present, either chirp type produces the same range measurement \eqref{eq:r_basic}, since the negatives in $\Delta f$ and $S$ cancel out for the down chirp.
However, in the case of a Doppler-induced shift $\Delta f_d$, the range measurement will be impacted in opposite ways.
This happens because the shift is constant for both chirps, whereas the slope of the modulation has a flipped sign.
\rev{The opposite shift is shown in Figure \ref{fig:normal_vs_doppler_scan}, where azimuths in the lower image alternate between up and down chirps creating a `zig-zag' effect in what should be flat features.
This is in contrast to the typically used sawtooth modulation, which consists only of up chirps and thus has a constant Doppler shift.
The opposite shift allows us to extract the Doppler velocity $u$, which is connected to $\Delta f_d$ through the transmitted signal frequency $f_t$ and wavelength $\lambda_t$ as
\begin{align}
    \Delta f_d = \f{2\,u\,f_t}{c} = \f{2\,u}{\lambda_t}.
\end{align}
}
The measured range from an up chirp with both shifts is
\begin{align}\label{eq:sys_2_eq_num1}
    r_\mathrm{up} = \frac{c \,\Delta f_t}{2 \, S} + \frac{c \,\Delta f_d}{2 \, S} = \tilde{r} + \Delta r_d,
\end{align}
where $\tilde{r}$ is the true range to the target and $\Delta r_d$ is the Doppler-induced range shift. For the down chirp, the range is
\begin{align}\label{eq:sys_2_eq_num2}
    r_\mathrm{down} = \frac{-c \Delta f_t}{-2\,S} + \frac{c \Delta f_d}{-2\,S} = \tilde{r} - \Delta r_d.
\end{align}
If a range measurement of the same target from both chirp types is recorded, we have a system of two equations and two unknowns. Solving this system for $\Delta r_d$, we can extract the corresponding relative velocity $u$, using $\beta = f_t/S$, as
\begin{subequations}
\begin{align}\label{eq:beta_intro}
    \Delta r_d &= \frac{c \,\Delta f_d}{2 \,S} = \frac{2 \,c \,u}{2 \, \lambda_t \,S}, \\
    u &= \frac{\Delta r_d \,S}{f_t} = \frac{\Delta r_d}{\beta}.
\end{align}
\end{subequations}

\begin{figure*}[t]
    \centering
    \begin{tikzpicture}
        % Main pipeline
        \node (pipeline) at (0,0) {\includegraphics[width=\textwidth]{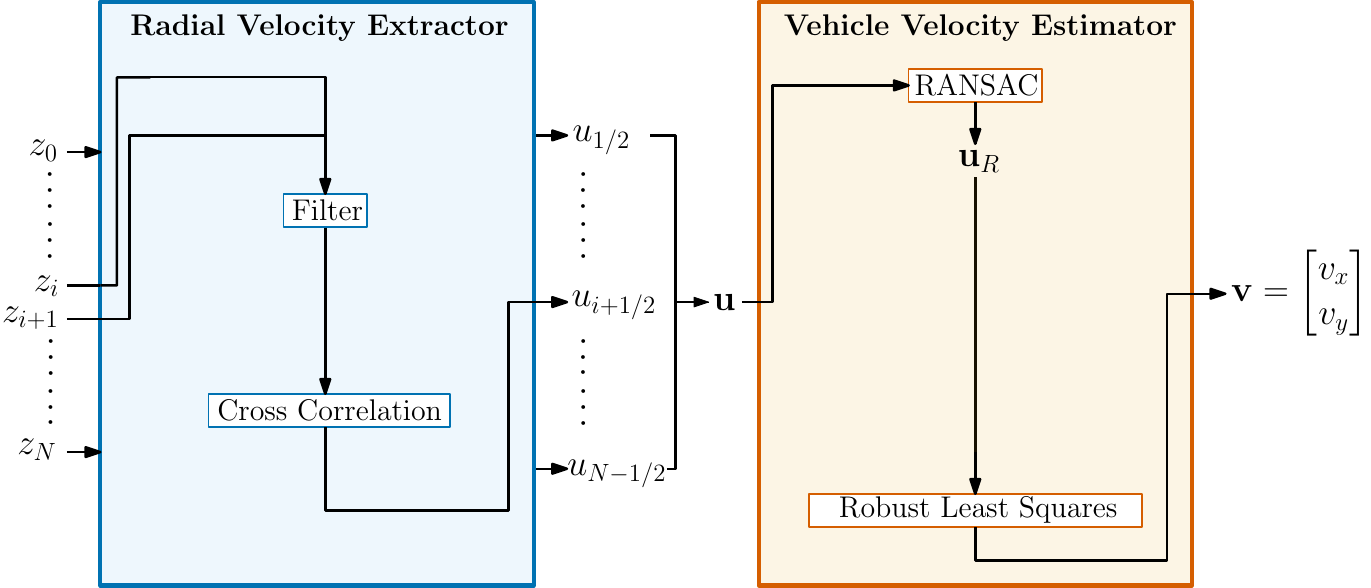}};
        % Raw signal
        \node[draw=black, inner sep=0pt] (raw_signal) at (-4.71,2.4) {\includegraphics[width=0.252\textwidth]{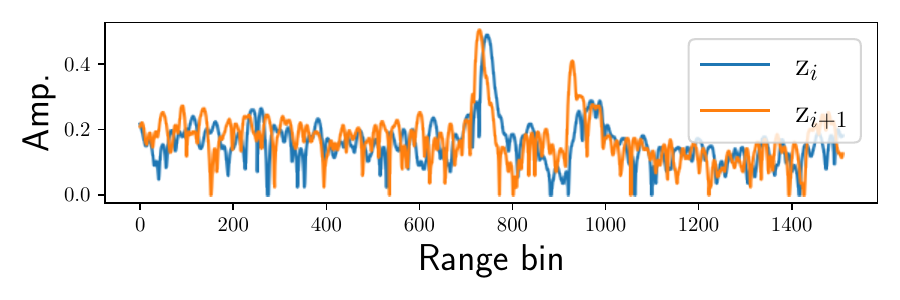}};
        % Cen signal
        \node[draw=black, inner sep=0pt] (raw_signal) at (-4.71,-0.15) {\includegraphics[width=0.252\textwidth]{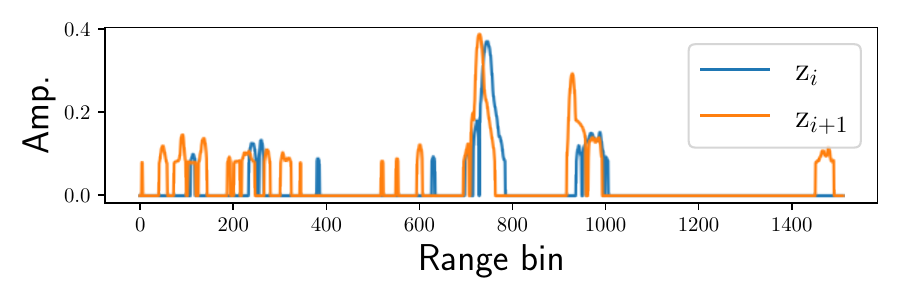}};
        % CC signal
        \node[draw=black, inner sep=0pt] (raw_signal) at (-4.71,-2.8) {\includegraphics[width=0.252\textwidth]{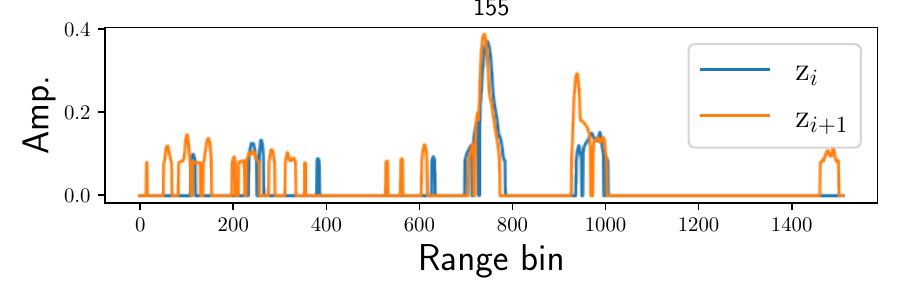}};
        % RANSAC
        \node[draw=black, inner sep=0pt] (raw_signal) at (3.8,-0.42) {\includegraphics[width=0.262\textwidth]{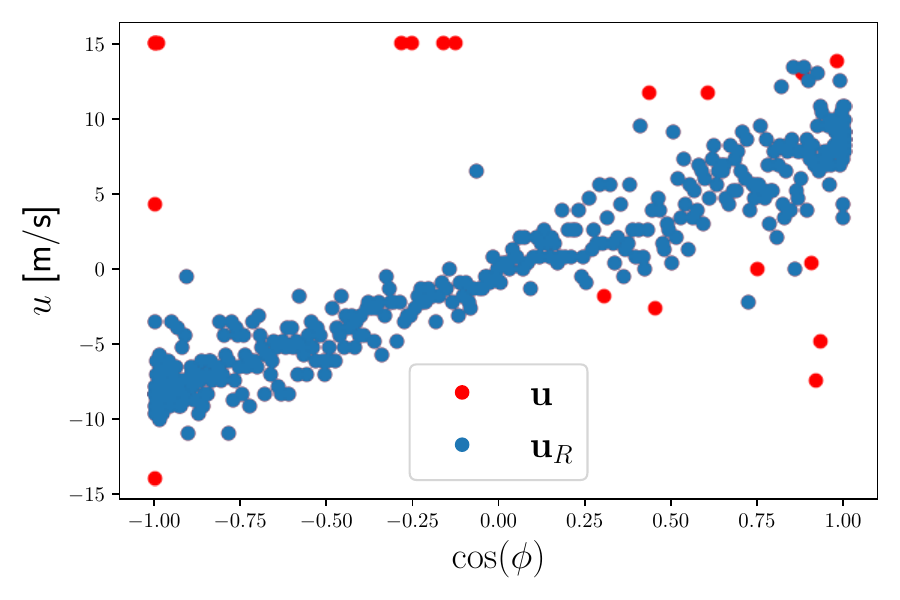}};
    \end{tikzpicture}
    
    \caption{\rev{The per-scan Doppler velocity estimation pipeline. First, consecutive pairs of raw radar intensity signals $\mathrm{z}_k, \, k \in \{0, N\}$ from azimuths $\phi_k$ are loaded into the radial velocity extractor. Each pair is filtered and a cross-correlation is run to estimate a radial velocity $u_j, \, j \in \{1/2, N-1/2\}$. Radial velocities for the entire scan $\mbf{u}$ are passed through RANSAC and the inliers $\mbf{u}_R$ used in a robust least-squares algorithm to estimate the ego-velocity $\mbf{v}$.}}
    \label{fig:pipeline}
\end{figure*}

\subsection{Doppler Velocity Extraction}
To extract the Doppler velocity measurement $u$ for each point, we need to find the Doppler-induced range offset $\Delta r_d$ for that point. This means obtaining range measurements from up and down chirp signals to use in \eqref{eq:sys_2_eq_num1} and \eqref{eq:sys_2_eq_num2}. In our case, where azimuths alternate in chirp type, this would involve identifying points from consecutive azimuths that correspond to the same target.
An alternative, entirely association-free way to find $\Delta r_d$ is to instead consider the entire azimuth at once. If we assume that consecutive azimuths are looking at mostly the same targets, then consecutive return signals should be oppositely shifted by the same $\Delta r_d$. Thus, a cross-correlation between consecutive signals would yield a shift of $2 \, \Delta r_d$.
The measurement is then formed without needing to ever extract or match explicit points in the return signals. For radar systems that produces output as a continuous stream, such as the Navtech RAS6 radar, measurements can be formed at the same time as individual azimuths arrive. \rev{The assumption that consecutive azimuths capture roughly the same targets becomes less accurate at further ranges, as the radar signal spreads out more. As a result, the signals are trimmed to a tuned maximum distance before being passed into the extractor. We nonetheless find that most of the signal ($200 \, \si{\meter}$ out of the available $250 \, \si{\meter}$) is needed for accurate alignment.} 

The cross-correlation on the raw signals is very noisy and error-prone. Thus, a signal filtering step is first taken. We find that a modified version of the filtering approach previously applied to radar pointcloud extraction in \cite{Cen_filtering} is highly effective in pre-processing the signals. Our filtering approach for each signal is as follows:
\begin{enumerate}
    \item Subtract the mean.
    \item Estimate the variance of the signal noise $\sigma^2_s$ (see \cite{Cen_filtering}).
    \item Apply a Gaussian filter ($\sigma_g = 15$).
    \item Re-weight the smoothed signal by the probability that each bin does not correspond to noise (see \cite{Cen_filtering}).
    \item Zero out signal values less than $2.5 \, \sigma_s$. 
\end{enumerate}
After filtering both azimuths, a normalized cross-correlation between the signals is performed to produce $\Delta r_d$ and then $u$. The velocity is assigned to the in-between azimuth value of the two azimuths. As we compute one radial velocity measurement for each unique pair of $N$ consecutive azimuths, we acquire $N-1$ measurements. This is in contrast to previous spinning Doppler radar works \cite{Vivet_Checchin_Chapuis_2013} and \cite{Rennie_Williams_Newman_DeMartini_2023}, which halved the number of used azimuths. This process is shown in the radial velocity extractor block of Figure \ref{fig:pipeline}.

\begin{figure}[t]
    \begin{tikzpicture}
        % Static subfigure
        \node (overlay) at (0.0,0) {\includegraphics[width=0.43\textwidth]{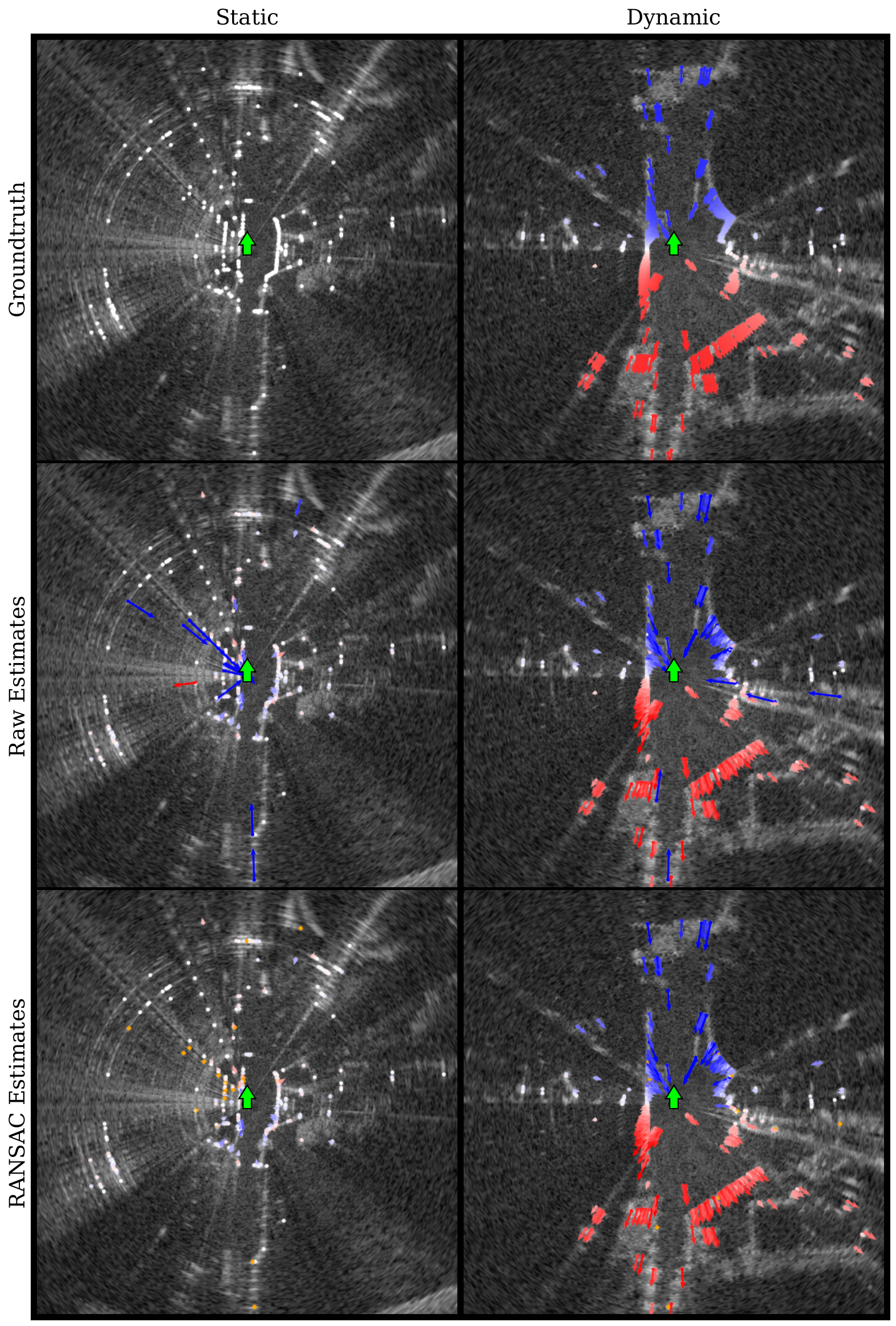}};
        \node (colourbar) at (0.1,-6.40) {\includegraphics[height=0.4\textwidth, angle=-90]{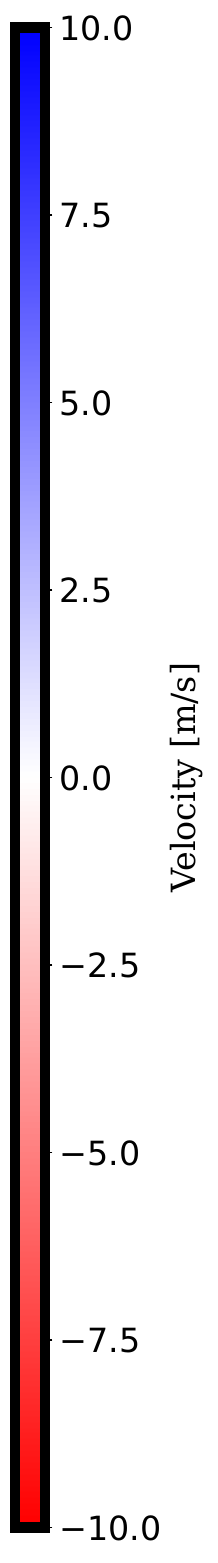}};
    \end{tikzpicture}
    \caption{\rev{Groundtruth and estimated radial velocities projected onto a radar pointcloud in a static and dynamic scene. RANSAC-rejected velocities are coloured in orange. The vehicle orientation is shown as a green arrow.}}
    \label{fig:proj_az_vel}
\end{figure}

\subsection{Velocity Pseudo-Measurement Estimator}\label{sec:vel_estimator}
In general, the Doppler velocity extraction is prone to outliers and produces velocities with a relatively high distribution about the true value. \rev{To improve robustness, we combine all radial velocity measurements for a single scan in an estimator to produce a per-scan ego-velocity pseudo-measurement}. This is visualized in the vehicle velocity estimator block of Figure \ref{fig:pipeline}.

First, we run RANSAC with a relatively loose outlier rejection threshold ($6 \, \si{\meter/\second}$) to filter out erroneous cross-correlation results and azimuths that correspond to dynamic objects. In environments where very few environmental points are present, it is easy for RANSAC to instead latch onto the dynamic objects in the scene. To overcome this issue, we include a loose prior to the RANSAC model. This prior discards RANSAC candidate solutions that deviate too far, more than $6 \, \si{\meter/\second}$ specifically, from the previous ego-velocity estimate, even if that candidate has a large number of inliers. We find that this is sufficient to mitigate the `capture' of the velocity estimate by a large number of dynamic objects in the scene, without restricting our estimator. \rev{This is reminiscent of the approaches taken in \cite{Michaelis_Berthold_Luettel_Wuensche, 10610311}.}

After RANSAC, the inlier azimuth velocities are fed to a robust least-squares (LS) algorithm. The LS algorithm estimates the forward $v_x$ and side $v_y$ ego-velocities as
\begin{align}
    \mbf{v} = \bbm v_x \\ v_y \ebm.
\end{align}
Given a vector of Doppler velocity measurements $\mbf{u}$ corresponding to azimuths $\mbs{\phi}$, the algorithm iteratively minimizes the objective function $J(\mbf{v})$:
\begin{align}
    J(\mbf{v}) &= \frac{1}{2} \mbf{e}(\mbf{v})^\trans \mbf{W} \mbf{e}(\mbf{v}),\\
    \mbf{e}(\mbf{v}) &= \bbm \cos{\mbs{\phi}} & \sin{\mbs{\phi}} \ebm \mbf{v} - \mbf{u},\\
    \mbf{W} &= \diag\left(\frac{1}{1 + (\mbf{e}(\mbf{v})/\rho)^2}\right),
\end{align}
where $\cos$ and $\sin$ act element-wise on $\mbs{\phi}$ and $\mbf{W}$ corresponds to a Cauchy weight with parameter $\rho$, which we set to $0.8$. \rev{The same approach is taken in \cite{Kellner_2013, Kellner_2014}, except with per-point radial velocities instead of our inter-azimuth ones.} It was found that a velocity-dependent bias existed in the ego-velocity estimates. \rev{This bias is potentially caused by the movement of the vehicle impacting the rotation of the radar disk, with more aggressive motions having a stronger impact. A velocity-dependent linear model was fit to holdout data and used to subtract off the bias during experiments.}

This approach generates a single ego-velocity estimate per radar scan, to which we assign a timestamp corresponding to the middle of the scan.
In the future, we hope to implement a continuous-time estimator to properly account for the timestamp at which each Doppler velocity is received.
Since most radars return only a few hundred azimuths per $360\si{\degree}$ sweep (400 for the Navtech RAS6 radar), this estimation is extremely fast. Although we find that using a Cauchy loss is necessary for best performance, \rev{adequate results may be produced without a robust cost function} depending on the requirements of the application. In this case, the estimation problem becomes linear in $\mbf{v}$ and can be solved in one step.

\rev{Figure \ref{fig:proj_az_vel} shows a qualitative result of the pipeline, with individually estimated inter-azimuth radial velocities projected onto points extracted using the BFAR extractor \cite{bfar} in the left-nearest azimuth. The pseudo-velocity measurement is not used.} The groundtruth radial velocities are generated by projecting the groundtruth vehicle ego-velocity onto each azimuth. It can be seen that the raw extracted radial velocities have obvious outliers, which are removed using RANSAC.
\section{Experiments}
    \label{sec:experiments}
    \begin{figure}[t]
    \centering

    \begin{tikzpicture}
        % Suburb
        \node[rotate=0] at (0.0,3.0) {{Suburbs}};
        \node (glen_radar) at (0.0,0) {
        \includegraphics[width=0.245\linewidth]{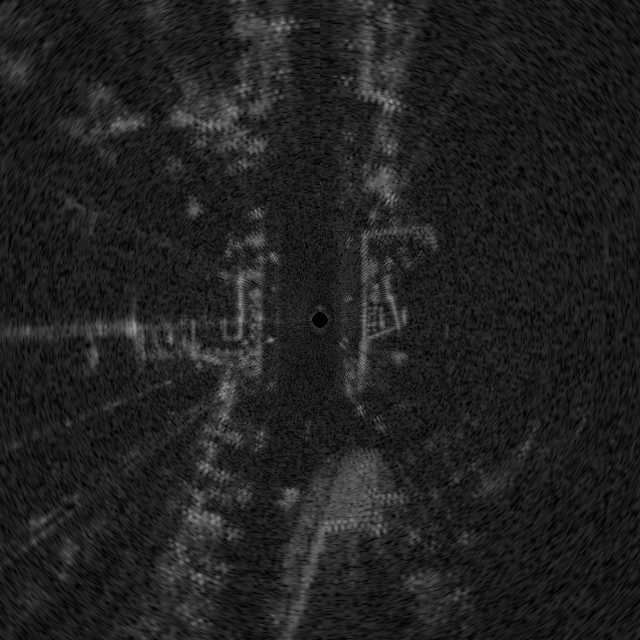}
        };
        \node (glen_camera) at (0.0,1.9) {
        \includegraphics[width=0.245\linewidth]{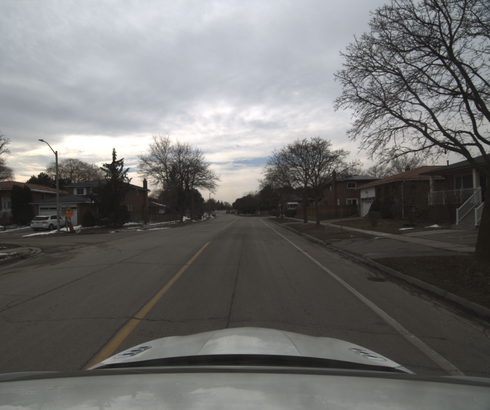}
        };
        
        % Highway
        \node[rotate=0] at (0.245\linewidth,3.0) {{Highway}};
        \node (hyw7_radar) at (0.245\linewidth,0) {
        \includegraphics[width=0.245\linewidth]{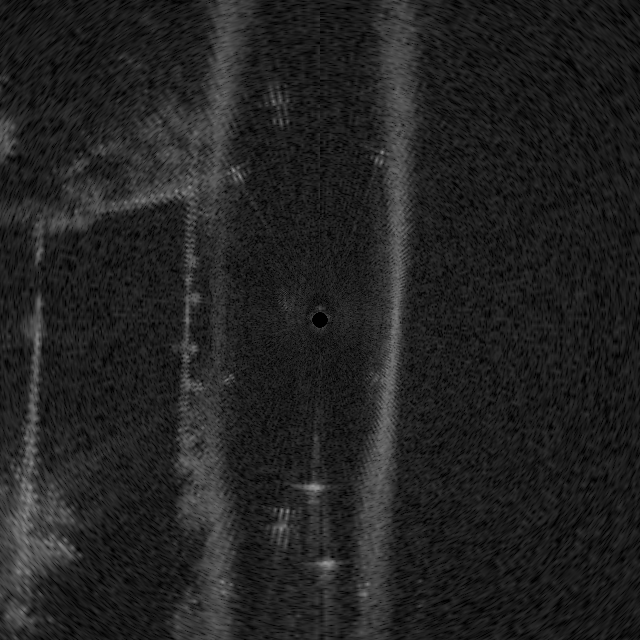}
        };
        \node (hwy7_camera) at (0.245\linewidth,1.9) {
        \includegraphics[width=0.245\linewidth]{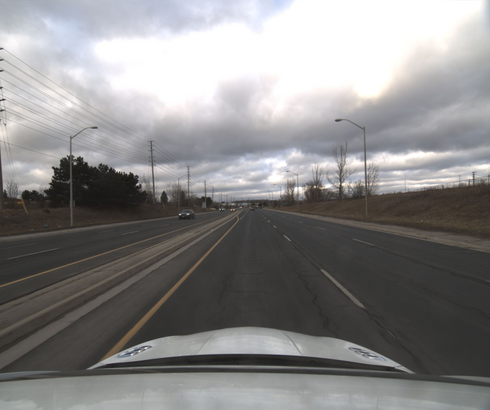}
        };
        % Tunnel
        \node[rotate=0] at (2*0.245\linewidth,3.0) {{Tunnel}};
        \node (tunnel_radar) at (2*0.245\linewidth,0) {
        \includegraphics[width=0.245\linewidth]{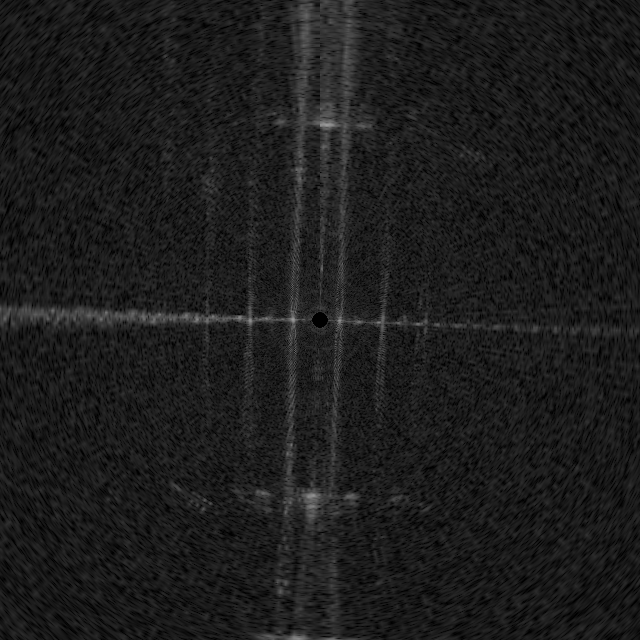}
        };
        \node (tunnel_camera) at (2*0.245\linewidth,1.9) {
        \includegraphics[width=0.245\linewidth]{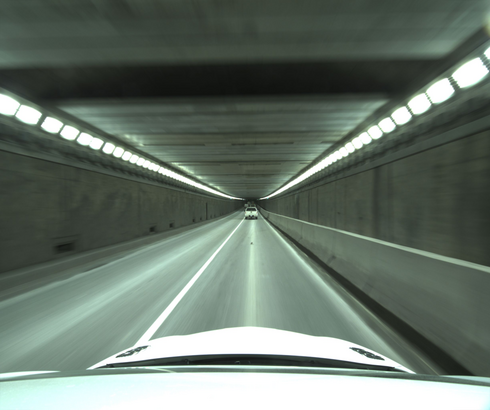}
        };
        % Skyway
        \node[rotate=0] at (3*0.245\linewidth,3.0) {{Skyway}};
        \node (skyway_radar) at (3*0.245\linewidth,0) {
        \includegraphics[width=0.245\linewidth]{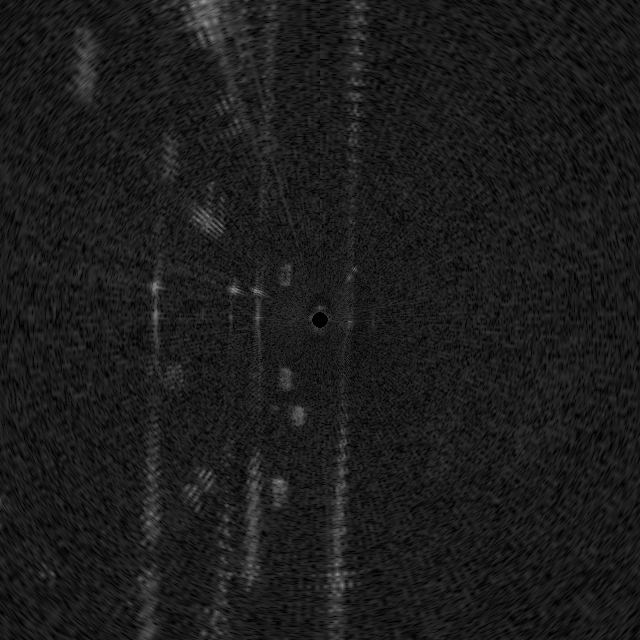}
        };
        \node (skyway_camera) at (3*0.245\linewidth,1.9) {
        \includegraphics[width=0.245\linewidth]{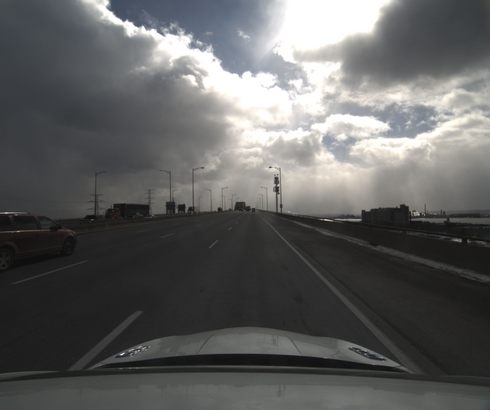}
        };
        
        \node[rotate=0, red, overlay] at (1.5*0.245\linewidth,-1.5) {\makecell[c]{Increasing difficulty}};
        \draw[-{Triangle[length=1.8mm,width=2.8mm]}, overlay, red, line width=1.5pt](2.2,-1.25) -- (4.2,-1.25);
    \end{tikzpicture}
    \vspace{0.02cm}
    \caption{\rev{Camera and radar data from the four trajectory types. Odometry is more challenging left to right due to fewer distinct geometric features.}}
    \label{fig:traj_vis}
\end{figure}

\subsection{Considered Pipelines}\label{sec:pipelines}
\subsubsection{ICP Baseline}
\rev{The two radar baseline pipelines (`B1' and `B2' in Table~\ref{tab:odom_results}) are based on the Teach and Repeat localization framework \cite{VTR} used in \cite{are_we_ready_for}. BFAR-extracted radar pointclouds are fed to p2p ICP in combination with a WNOA prior to produce an odometry estimate. The $SE(2)$ pose and $\mbs{\varpi} = \bbm v_x & v_y & \omega \ebm^\trans$ ego-velocity is estimated in continuous time. The ego-velocity estimate is used to compensate for the Doppler-based range offset in the pointclouds. Please refer to \cite{are_we_ready_for} for full details. B2 additionally adds in a preintegrated heading gyroscope factor between consecutive radar frames. The continuous-pose estimate is queried at the start and end timesteps of the preintegrated factor, and the estimated change in orientation is contrasted to the preintegrated value. We only use a gyroscope, instead of a full IMU, to avoid estimating the gravity vector and to highlight the benefit from adding a simple, single-axis sensor. We leave the integration of a full IMU, including estimating relevant biases, to future work. These pipelines are relatively slow due to ICP.

We also include a lidar ICP baseline (`B3'). This pipeline, also used in \cite{are_we_ready_for}, is set up in the same way as `B1', but estimates a full $SE(3)$ pose and 6D velocity. It runs well below real-time due to the use of continuous-time ICP with several thousand points (maximum $10,000$ for our experiments), but serves as a guideline for best performance that could be achieved with a more information-dense sensor.}

\subsubsection{Doppler Radar ICP}
Two new spinning radar odometry pipelines directly extend the baselines. These pipelines (`N1' and `N2' in Table \ref{tab:odom_results}) add a Doppler measurement-based term to the optimization problem. As discussed in Section~\ref{sec:vel_estimator}, the Doppler measurement estimator is used to estimate a forward and side ego-velocity $\mbf{v}$ for the timestamp at the midpoint of each scan. This estimate is then included as an additional cost term in the radar ICP pipelines by querying the continuous-velocity estimate at the scan midpoint timestamp and contrasting it with $\mbf{v}$. A fixed noise covariance is tuned for this loss term. The added term is used both without (N1) and with (N2) a preintegrated heading gyroscope factor. These pipelines are relatively slow due to ICP.

\subsubsection{Doppler Radar + Gyroscope}
\rev{The final novel spinning radar odometry pipeline (`N3' in Table \ref{tab:odom_results}) is adapted from~\cite{need_for_speed} and directly integrates Doppler-based linear velocity pseudo-measurements $\mbf{v}$ and preintegrated heading gyroscope factors $\Delta \mbf{C}$ to estimate a pose. This is done by propagating the odometry estimate by a pose change term between the previous and current radar frames separated by $\delta t \, \si{\second}$, with a position change of $\delta t \, \mbf{v}$ and an orientation change of $\Delta \mbf{C}$. This pipeline is extremely fast, but is highly dependent on the quality of the gyroscope. We do not estimate the gyroscope bias in any pipeline for a fair comparison with N3, where the bias is unobservable.}

\begin{figure*}[t]
    \centering
    \includegraphics[width=\linewidth]{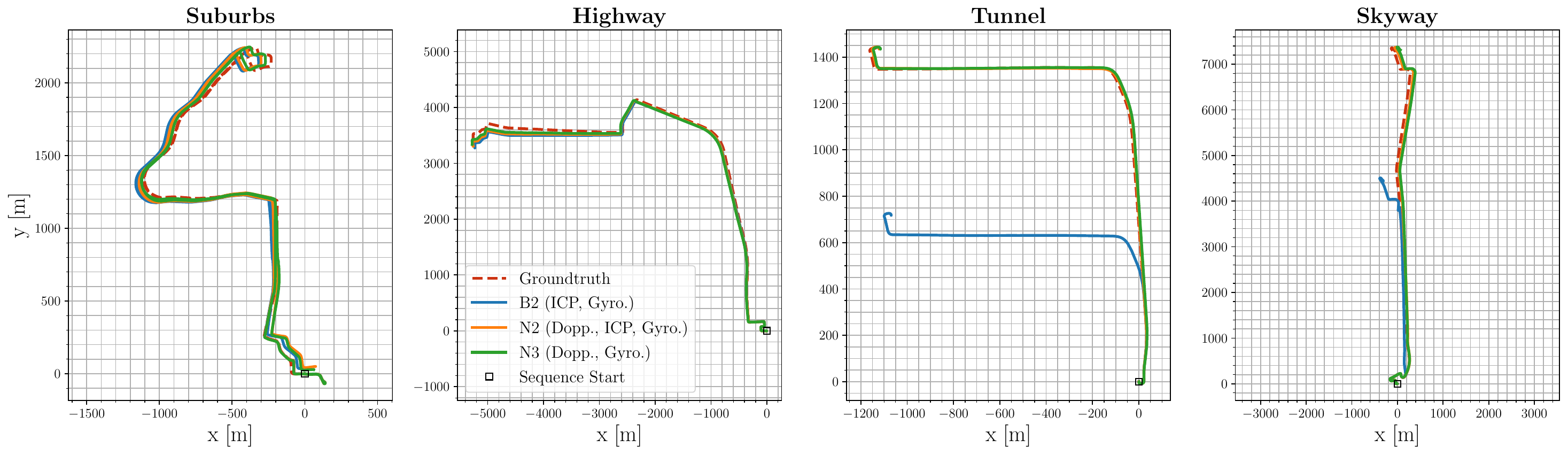}
    \begin{tikzpicture}[overlay]
        \node[rotate=0, red] at (0,0.25) {\makecell[c]{Increasing difficulty}};
        \draw[-{Triangle[length=1.8mm,width=2.8mm]}, red, line width=1.5pt](-1,0.5) -- (1,0.5);
    \end{tikzpicture}
    \caption{Visualization of odometry performance for one sequence from each of the four types of trajectories. The baseline radar ICP with gyroscope data algorithm (B2) is shown alongside the two new algorithms (N2, N3) making use of spinning radar Doppler measurements. Groundtruth is shown as a red dotted line. The trajectories are presented in increasing order of difficulty from left to right. The new methods perform on par with the baseline in the two easiest cases, suburbs and highway, and continue to provide valid odometry results in the two harder cases, tunnel and skyway, where the baseline fails.}
    \label{fig:traj_plots}
\end{figure*}

\subsection{Dataset}
We used the Boreas platform \cite{boreas} to collect Doppler-enabled spinning radar data in progressively more challenging environments, as visualized in Figure \ref{fig:traj_vis}. Boreas is equiped with a Navtech RAS6 radar, which operates between $76$ and $77\, \si{\giga\hertz}$, has a range resolution of $4.38 \, \si{\cm}$, spins at $ 4\, \si{\hertz}$, and generates 400 azimuth measurements per rotation. \rev{The sensor has a maximum range of $250 \, \si{\meter}$, but only returns from the first $200 \, \si{\meter}$ and $80 \, \si{\meter}$ are used in the radial velocity and pointcloud extractors, respectively. To estimate the $\beta$ parameter from \eqref{eq:beta_intro}, we approximate $f_t$ as the midpoint frequency of the ramp and $S$ as the change in frequency of a chirp divided by the time it takes to complete a chirp. As this is an approximation, we find that a $0.944$ correction factor on $\beta$ is needed to get the extracted velocities to better match the tuning data.} The signal has a beamwidth of $1.8\si{\degree}$, meaning that consecutive azimuths overlap by $0.9\si{\degree}$ and letting us assume that consecutive signals have returns from the same targets. The groundtruth is generated from post-processed GNSS, IMU, and wheel encoder data using Applanix's RTX POSPac software. A caveat is that the gyroscope used for generating the groundtruth is the same one that is used in experiments, as it was the only gyroscope on the platform.\footnote{We plan to integrate a separate heading gyroscope in the future.} We use the unprocessed data in our algorithms, but it is nonetheless lightly correlated with the groundtruth.

\renewcommand{\arraystretch}{1.25}
\begin{table}[t!]
    \centering
    \caption{\rev{Percent 2D translational odometry drift and runtime. Best radar results are bolded.}}
    \label{tab:odom_results}
    \scalebox{0.92}{
    \begin{tabular}{?{1.2pt} c ?{1.2pt} c ?{1.2pt} c | c | c ?{1.2pt} c | c ?{1.2pt} c ?{1.2pt}}
    \Cline{1.2pt}{3-8}
        \multicolumn{2}{c?{1.2pt}}{} & \multicolumn{3}{c?{1.2pt}}{\textbf{New}} & \multicolumn{3}{c?{1.2pt}}{\textbf{Baselines}} \\
        \cline{3-8}
         \multicolumn{2}{c?{1.2pt}}{}  & N1 & N2 & N3 & B1 & B2 & B3 \\
        \Cline{1.2pt}{2-8}
         \multicolumn{1}{c?{1.2pt}}{} & Sensor & Radar & Radar & Radar & Radar & Radar & Lidar \\
        \cline{2-8}
         \multicolumn{1}{c?{1.2pt}}{} & ICP & \checkmark & \checkmark &  & \checkmark & \checkmark & \checkmark \\
        \cline{2-8}
         \multicolumn{1}{c?{1.2pt}}{} & Gyro. &  & \checkmark & \checkmark &  & \checkmark &  \\
         \cline{2-8}
         \multicolumn{1}{c?{1.2pt}}{} & Doppler & \checkmark & \checkmark & \checkmark &  & &  \\
    \Cline{1.2pt}{1-8}
    \multicolumn{2}{?{1.2pt}c?{1.2pt}}{Dopp. Extr. [$\si{\milli\second}$]} & 45 & 45 & 45 & {$\mbs{\sim}$} & $\mbs{\sim}$ & {$\mbs{\sim}$} \\
    \hline
    \multicolumn{2}{?{1.2pt}c?{1.2pt}}{Odometry [$\si{\milli\second}$]} & 110 & 110 & \textbf{1} & 136 & 113 & 415 \\
    \hline
    \multicolumn{2}{?{1.2pt}c?{1.2pt}}{Total FPS} & 6.5 & 6.5 & \textbf{21.7} & 7.4 & 8.8 & 2.4 \\
    \Cline{1.2pt}{1-8}
        \multirow{5}{*}{\rotatebox[origin=c]{90}{Suburbs}} 
        & 1 & {1.10} & \textbf{0.58} & 1.39 & 1.23 & 0.59 & 0.17 \\
    \cline{2-8}
         & 2 & 0.79 & \textbf{0.44} & {0.64} & 1.08 & 0.58 & 0.13 \\
    \cline{2-8}
         & 3 & 1.13 & \textbf{0.51} & 1.42 & 1.18 & {0.54} & 0.15 \\
    \cline{2-8}
         & 4 & 1.25 & \textbf{0.42} & {0.63} & 1.08 & \textbf{0.42} & 0.13 \\
    \Cline{1.2pt}{2-8}
         & {\color{blue}Avg. [\%]} & {\color{blue} 1.07} & {\color{blue} \textbf{0.49}} & {\color{blue}{1.02}} & {\color{blue}1.14} & {\color{blue}0.53} & {\color{blue}0.15} \\
    \Cline{1.2pt}{1-8}
        \multirow{5}{*}{\rotatebox[origin=c]{90}{Highway}}
        & 1 & 1.69 & \textbf{0.67} & {1.17} & 1.91 & 0.91 & 0.48 \\
    \cline{2-8}
         & 2 & 1.87 & \textbf{0.57} & {1.17} & 2.07 & 0.62 & 0.20 \\
    \cline{2-8}
         & 3 & 1.90 & \textbf{0.50} & {1.60} & 2.00 & 0.63 & 0.22 \\
    \cline{2-8}
         & 4 & 1.33 & \textbf{0.40} & {1.29} & 1.54 & 0.44 & 0.17 \\
    \Cline{1.2pt}{2-8}
         & {\color{blue}Avg. [\%]} & {\color{blue} 1.70} & {\color{blue} \textbf{0.54}} & {\color{blue}{1.31}} & {\color{blue}1.88} & {\color{blue}0.65} & {\color{blue}0.27} \\
    \Cline{1.2pt}{1-8}
        \multirow{5}{*}{\rotatebox[origin=c]{90}{Tunnel}}
        & 1 & 3.86 & \textbf{0.56} & {0.75} & 29.80 & 29.43 & 26.88 \\
    \cline{2-8}
         & 2 & 2.89 & \textbf{0.45} & {0.70} & 32.06 & 31.75 & 29.11 \\
    \cline{2-8}
         & 3 & 5.46 & \textbf{0.71} & {0.99} & 34.25 & 33.83 & 21.62 \\
    \cline{2-8}
         & 4 & 3.37 & \textbf{0.49} & {0.80} & 30.99 & 29.82 & 29.46 \\
    \Cline{1.2pt}{2-8}
         & {\color{blue}Avg. [\%]} & {\color{blue}3.90} & {\color{blue} \textbf{0.55}} & {\color{blue}{0.81}} & {\color{blue}31.77} & {\color{blue}31.21} & {\color{blue}26.77} \\
    \Cline{1.2pt}{1-8}
        \multirow{5}{*}{\rotatebox[origin=c]{90}{Skyway}}
        & 1 & 4.05 & \textbf{0.39} & {0.64} & 36.58 & 31.93 & 11.35 \\
    \cline{2-8}
         & 2 & 2.44 & \textbf{0.44} & {0.64} & 3.14 & 0.83 & 17.33 \\
    \cline{2-8}
         & 3 & 7.14 & \textbf{0.70} & {0.73} & 58.17 & 58.47 & 34.62 \\
    \cline{2-8}
         & 4 & 5.58 & \textbf{0.58} & {0.76} & 56.77 & 55.19 & 20.49 \\
    \Cline{1.2pt}{2-8}
         & {\color{blue}Avg. [\%]} & {\color{blue}4.80} & {\color{blue} \textbf{0.53}} & {\color{blue}{0.69}} & {\color{blue}38.66} & {\color{blue}36.60} & {\color{blue}20.95} \\
    \Cline{1.2pt}{1-8}
    \end{tabular}
    }
\end{table}

We collected four repeated runs on four progressively more difficult sequences totaling approximately $111 \, \si{\km}$. Parameters were tuned on an additional 6 holdout sequences from a mix of sequence types. The sequence types are:
\begin{enumerate}
    \item Suburbs (easy): These $\approx 7.9 \,\si{\km}$ sequences follow the suburban Glen Shields route from \cite{boreas} with speed limits of $30-60 \, \si{\km/\hour}$. All parts of the trajectory have multiple visible and unique geometric features.
    \item Highway (medium): \rev{These $\approx 9.1 \,\si{\km}$ sequences start in a suburban environment and then exit onto a highway with speed limits up to $80 \, \si{\km/\hour}$. Parts of the highway are geometrically challenging with few static features.}
    \item Tunnel (hard): These $\approx 2.5 \, \si{\km}$ sequences drive through the $840 \, \si{\meter}$ long Thorold Tunnel in St. Catherines, Ontario, before turning onto a country road. The speed limit is up to $80 \, \si{\km/\hour}$. The inside of the tunnel is consistently geometrically degenerate, with only two continuous, parallel walls visible.
    \item Skyway (hard): \rev{These $\approx 8.3 \,\si{\km}$ sequences traverse the $2,560 \, \si{\meter}$ long Burlington Bay James N. Allan Skyway in Burlington, Ontario, with speed limits of up to $100 \, \si{\km/\hour}$. The skyway has very few visible static features, but many vehicles in front and behind the car. The static features that do exist are repetitive.}
\end{enumerate}

\subsection{Odometry Results}
The translational odometry drift from all pipelines is presented in Table \ref{tab:odom_results}. The drift is reported as an average value of the KITTI-style odometry metrics \cite{kitti} from subsequences of length (100, 200, \dots, 800). We omit rotational drift values for brevity and because they show similar trends. A visualization of the odometry is shown in Figure~\ref{fig:traj_plots}.

It can be seen that ICP-based methods generally drop in performance as the geometric difficulty of the sequences increases. \rev{For all three baselines, including the lidar one,} the estimation pipeline fails entirely as soon as the car enters the tunnel or skyway. This is visualized in the velocity estimates shown in Figure \ref{fig:boreas}. All new pipelines that incorporate Doppler measurements produce functional odometry results regardless of the type of sequence. Using all available sources of information (N2) yields the best radar performance in all sequences and is the only radar pipeline that maintains excellent performance ($\approx 0.5\%$ average drift) without dependence on the sequence type. \rev{N2 is still two and three times worse than the lidar-only pipeline B3 in the suburbs and highway respectively, but maintains the same high level of performance when B3 fails.}

Despite losing out in performance to N2, directly integrating Doppler and gyroscope measurements (N3) performs well in all cases. This highlights the value that a heading gyroscope brings to an estimation pipeline, which we have similarly noted in our previous work \cite{Burnett_Schoellig_Barfoot_2024, need_for_speed, imu_as_input}. This is further supported by the root mean square error (RMSE) of the extracted Doppler velocities shown in Table \ref{tab:dopp_extr_results}. Though the RMSE is lower in the suburbs and highway, direct integration performs the best in the tunnel and skyway, where there are few turns and the gyroscope is very reliable.

When gyroscope data is not used (B1 and N1), the extracted Doppler measurements noticeably improve results in 15 out of 16 sequences and prevent complete odometry failure in the tunnel and skyway. This result is achieved with only a slight computational costs needed to extract the measurements, but notably without any hardware changes.

Overall, our results draw parallels to the findings for automotive phased-array radar \cite{Kubelka_Fritz_Magnusson_2023} and FMCW lidar \cite{picking_up_speed}. Similarly to \cite{picking_up_speed}, we find that adding Doppler measurements to an ICP pipeline results in modest improvement in geometrically well-defined scenes and prevents failure in geometrically degenerate environments. Compared to \cite{Kubelka_Fritz_Magnusson_2023}, we also find that directly integrating Doppler measurements with an inertial measurement sensor (IMU for them and heading gyroscope for us) performs remarkably well in all environments. However, they find that direct integration outperforms ICP-based methods in degenerate conditions, even when Doppler and IMU data is used. We find that using all available sources of information outperforms direct integration in all sequences. This difference may be due to \cite{Kubelka_Fritz_Magnusson_2023} using an $80 \si{\degree}$ FOV automotive phased-array radar, which makes it more likely to only observe geometrically degenerate features as compared to a $360 \si{\degree}$ spinning radar.

Table \ref{tab:odom_results} also shows timing information based on single-thread performance of a Lenovo P1 Laptop with Intel(R) Core(TM) i7-12800H CPU @ $4.8 \, \si{\giga\hertz}$ and $32 \, \si{\giga\byte}$ of RAM. We list the Doppler extraction time, meaning the amount of time required to generate a Doppler pseudo-measurement according Figure \ref{fig:pipeline}, separately, as this can be theoretically run on the radar itself. The majority of this runtime comes from filtering each signal and computing a cross-correlation between neighbours, both of which could be done as azimuths arrive. All radar pipelines are real-time in the sense that they run faster than frames arrive ($\geq 4 \, \si{\hertz}$).

\renewcommand{\arraystretch}{1.25}
\begin{table}[t!]
    \centering
    \caption{Extracted Doppler velocity RMSE with mean in brackets.}
    \label{tab:dopp_extr_results}
    \begin{tabular}{?{1.2pt} c ?{1.2pt} c | c | c | c ?{1.2pt}}
    \Cline{1.2pt}{2-5}
        \multicolumn{1}{c?{1.2pt}}{} & Suburbs & Highway & Tunnel & Skyway \\
        \cline{2-5}\cline{2-5}
    \Cline{1.2pt}{1-5}
         $v_x \; [\si{\meter/\second}]$ & 0.13 (0.01) & 0.18 (0.02) & 0.19 (-0.02) & 0.18 (0.01) \\
         \hline
         $v_y \; [\si{\meter/\second}]$ & 0.12 (0.01) & 0.17 (0.00) & 0.21 (-0.05) & 0.27 (0.05) \\
    \Cline{1.2pt}{1-5}
    \end{tabular}
\end{table}
\section{Conclusion}
    \label{sec:conclusion}
    We present a method to analytically extract \rev{spinning radar} Doppler velocity measurements and make use of them in modern radar odometry pipelines. We test our approach in four progressively more challenging environments: suburbs, highway, tunnel, and skyway. The use of Doppler measurements improves performance in easy geometric environments and maintains good performance in geometrically degenerate situations, where the baselines fail. We show this in ICP-based pipelines and in a fast ego-velocity estimation pipeline, where odometry is generated by directly integrating Doppler and heading velocities. \rev{When a fast solution is desired, the direct velocity estimation pipeline is a promising solution. When best performance is needed, an ICP-based pipeline with Doppler and gyroscope measurements produces highly accurate and consistent results in all road environments. The extraction of Doppler measurements is possible and beneficial in all types of environments, regardless of the weather, and at no hardware cost.}

Future extensions to the work presented here will include evaluating the impact of Doppler measurements on mapping, localization, and \rev{dynamic} object \rev{detection and} tracking. \rev{It would also be interesting to compare the performance of Doppler-enabled spinning radar to Doppler-enabled lidar.}

%if needed
\renewcommand*{\bibfont}{\footnotesize}
%--------
\printbibliography

\end{document}